\documentclass{article}

\usepackage{microtype}
\usepackage{graphicx}
\usepackage{subcaption}
\usepackage{booktabs} 
\usepackage{multirow}
\usepackage{hyperref}

\usepackage[preprint]{icml2026}

\usepackage{amsmath}
\usepackage{amssymb}
\usepackage{mathtools}
\usepackage{amsthm}
\usepackage{colortbl}
\usepackage{enumitem}
\usepackage[capitalize,noabbrev]{cleveref}

\theoremstyle{plain}

\theoremstyle{definition}

\theoremstyle{remark}

\usepackage{xcolor}
\definecolor{lightpurple}{HTML}{EDE3FE}
\usepackage[textsize=tiny]{todonotes}
\usepackage{tcolorbox}
\usepackage{listings}
\tcbuselibrary{breakable}
\newtcolorbox{promptbox}{
  colback=gray!5,
  colframe=gray!40,
  boxrule=0.5pt,
  arc=2pt,
  left=6pt,
  right=6pt,
  top=6pt,
  bottom=6pt,
  breakable,
  fontupper=\ttfamily\small
}
\lstdefinestyle{promptstyle}{
  basicstyle=\ttfamily\small,
  breaklines=true,
  columns=fullflexible,
  frame=single,
  keepspaces=true,
  showstringspaces=false,
  backgroundcolor=\color{gray!5},
}

\icmltitlerunning{FactGuard: Agentic Video Misinformation Detection via Reinforcement Learning}

\begin{document}

\twocolumn[
  \icmltitle{FactGuard: Agentic Video Misinformation Detection \\via Reinforcement Learning}



  \icmlsetsymbol{equal}{*}

  \begin{icmlauthorlist}
    \icmlauthor{Zehao Li}{equal,ict,ucas}
    \icmlauthor{Hongwei Yu}{equal,ustb}
    \icmlauthor{Hao Jiang}{ict,ucas}
    \icmlauthor{Qiang Sheng}{ict,ucas}
    \icmlauthor{Yilong Xu}{ict,ucas}\\
    \icmlauthor{Baolong Bi}{ict,ucas}
    \icmlauthor{Yang Li}{ict,ucas}
    \icmlauthor{Zhenlong Yuan}{ict,ucas}
    \icmlauthor{Yujun Cai}{uq}
    \icmlauthor{Zhaoqi Wang}{ict,ucas}
  \end{icmlauthorlist}

  \icmlaffiliation{ict}{Institute of Computing Technology, Chinese Academy of Sciences, Beijing, China}
  \icmlaffiliation{ucas}{University of Chinese Academy of Sciences, Beijing, China}
  \icmlaffiliation{ustb}{University of Science and Technology Beijing, Beijing, China}
  \icmlaffiliation{uq}{The University of Queensland, Brisbane, Australia}
  \icmlcorrespondingauthor{Hao Jiang}{jianghao@ict.ac.cn}
  \icmlcorrespondingauthor{Zehao Li}{lizehao23z@ict.ac.cn}

  \icmlkeywords{Machine Learning, ICML}

  \vskip 0.3in
]



\printAffiliationsAndNotice{}  

\begin{abstract}
Multimodal large language models (MLLMs) have substantially advanced video misinformation detection through unified multimodal reasoning, but they often rely on fixed-depth inference and place excessive trust in internally generated assumptions, particularly in scenarios where critical evidence is sparse, fragmented, or requires external verification. To address these limitations, we propose \textbf{FactGuard}, an agentic framework for video misinformation detection that formulates verification as an iterative reasoning process built upon MLLMs. FactGuard explicitly assesses task ambiguity and selectively invokes external tools to acquire critical evidence, enabling progressive refinement of reasoning trajectories. To further strengthen this capability, we introduce a two-stage training strategy that combines domain-specific agentic supervised fine-tuning with decision-aware reinforcement learning to optimize tool usage and calibrate risk-sensitive decision making. Extensive experiments on FakeSV, FakeTT, and FakeVV demonstrate FactGuard's state-of-the-art performance and validate its excellent robustness and generalization capacity. 
\end{abstract}

\section{Introduction}
\label{sec:int}
The rapid growth of large-scale online content-sharing platforms, such as TikTok, has accelerated the spread of information~\cite{bu2023combating}. This accessibility creates a fundamental asymmetry between rapid misinformation propagation and delayed manual verification, allowing misleading content to influence public discourse beyond the limits of human intervention. Consequently, post-hoc fact-checking becomes increasingly inadequate, highlighting the need for accurate and timely automated misinformation detection~\cite{sheng2025combating, tiktec}.

Among various content modalities, videos have emerged as a dominant and particularly challenging medium for misinformation dissemination due to their rich temporal dynamics and inherent multimodal complexity. In recent years, increasing attention has therefore been devoted to video misinformation detection, yielding encouraging progress. However, most existing approaches~\cite{fakesv, fakett, qi-etal-2023-two} rely on task-specific discriminative models and lack the general understanding and reasoning capabilities, which are required for handling various verification needs in open-world scenarios.
\begin{figure}[t]
  \centering
  \includegraphics[width=1\linewidth]{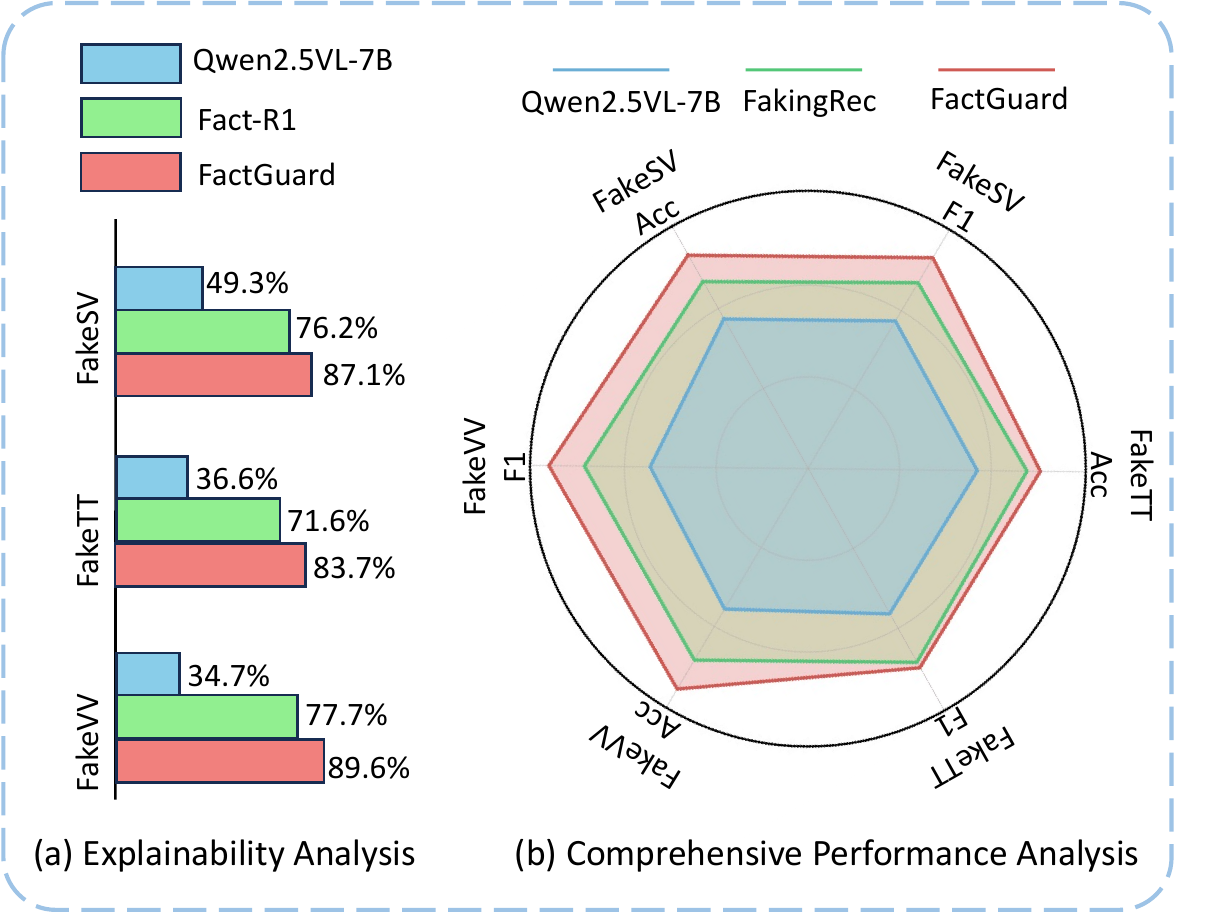}
\caption{
Comparison of video misinformation detection methods and our proposed \textbf{FactGuard} in terms of (a) explainability analysis and (b) comprehensive performance analysis.
}
\label{fig:sota}
\end{figure}

The rapid advancement of large-scale multimodal models has significantly improved multimodal content understanding and enabled reasoning-based approaches for misinformation verification. However, most existing methods remain constrained by a single-pass inference paradigm, lacking explicit mechanisms for uncertainty awareness and targeted evidence acquisition. In ambiguous or weakly verifiable scenarios, models therefore tend to rely on internally generated assumptions rather than grounding their reasoning in modality-specific or external evidence. As illustrated in Figure~\ref{fig:case_2}(a), this behavior results in \textbf{cross-modal hallucinations}, where the model fabricates or misattributes visual or textural evidence that is not present in the input, and consequently leads to \textbf{erroneous verification outcomes}, namely confident but incorrect judgments about the factuality of the underlying claim. Consequently, reliable misinformation verification requires a system that can (1) identify when available information is insufficient, (2) selectively acquire targeted external evidence, and (3) iteratively refine its judgments based on new observations.

Motivated by this insight, we propose \textbf{FactGuard}, an end-to-end agentic framework for video misinformation detection built on multimodal large language models (MLLMs). FactGuard formulates verification as an uncertainty-aware, iterative decision-making process, selectively invoking external tools when necessary to support reliable verification. Specifically, FactGuard introduces an agentic reasoning pipeline that integrates multimodal deliberation with explicit tool invocation and evidence-driven refinement. When the available information is insufficient for confident verification, the model adaptively selects external tools based on its reasoning state, including \textit{FactProbe} for external knowledge verification and \textit{ClipScout} for targeted visual evidence inspection, to supplement textual and visual inputs. The acquired evidence is incorporated into subsequent reasoning stages, enabling more informed and reliable verification outcomes. To further strengthen this capability, we construct a multimodal agentic Chain-of-Thought dataset for misinformation detection and perform supervised fine-tuning to establish structured reasoning and tool-invocation behaviors. We additionally incorporate a decision-aware reinforcement learning strategy that reinforces evidence-grounded reasoning while explicitly modeling tool usage, asymmetric error costs, and decision preferences under uncertainty. Together, these design choices enable FactGuard to move beyond existing approaches, establishing a general, and interpretable verification framework, as shown in Figure~\ref{fig:sota}.

Our main contributions are as follows:
\begin{itemize}[leftmargin=*]
    \item We propose \textbf{FactGuard}, an agentic multimodal framework that formulates video misinformation detection as an iterative verification process with self-reflective reasoning and selective evidence acquisition.

    \item We construct a multimodal agentic Chain-of-Thought dataset for misinformation detection and perform targeted CoT-based supervised fine-tuning to inject domain-specific reasoning behaviors into MLLMs.

    \item We develop a decision-aware reinforcement learning strategy that explicitly models tool usage, asymmetric error costs, and decision-making preferences, leading to more calibrated and reliable verification outcomes.

    \item Extensive experiments on FakeSV, FakeTT, and FakeVV datasets demonstrate that FactGuard achieves state-of-the-art performance and consistently outperforms existing methods by a significant margin.
\end{itemize}

\section{Related Work}
\label{sec:rel}
\subsection{Misinformation Detection}
\begin{figure*}[t]
  \centering
  \includegraphics[width=1\linewidth]{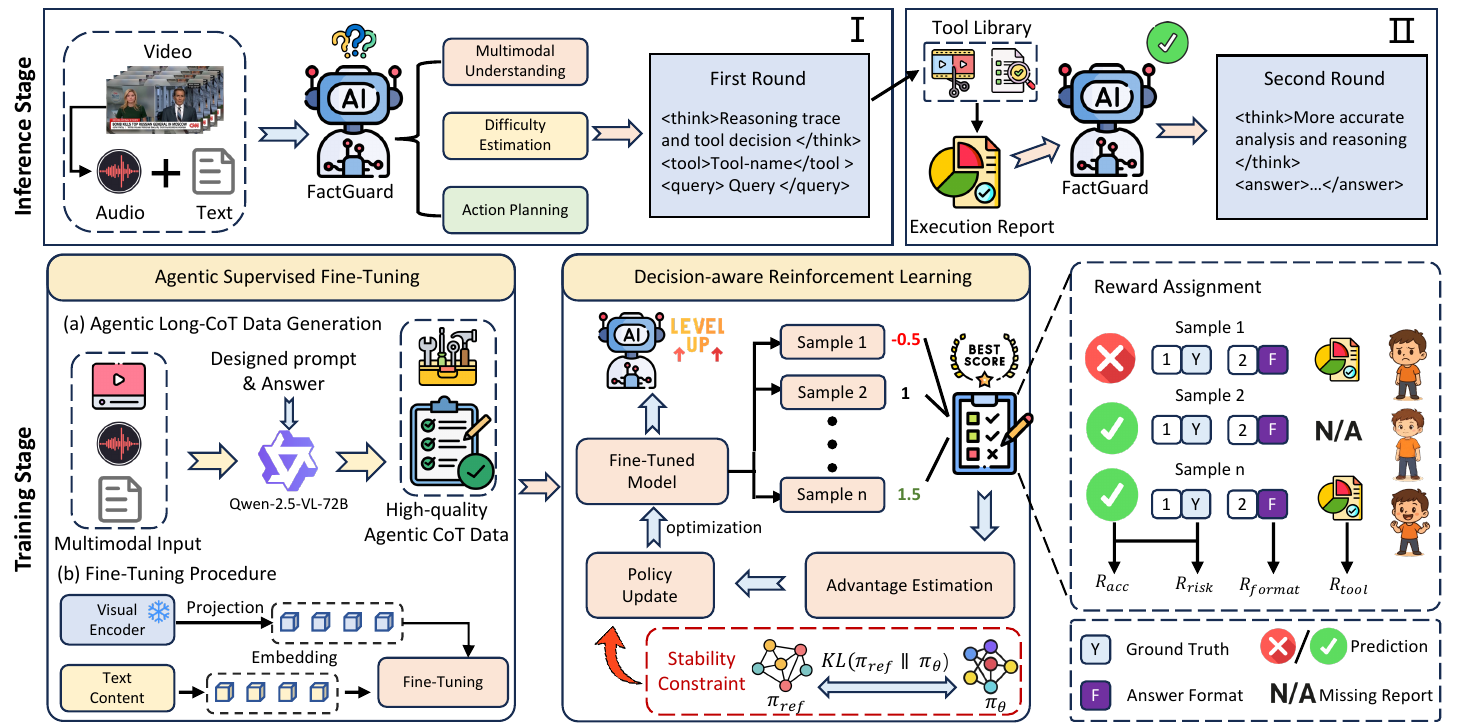}
\caption{
\textbf{Pipeline of FactGuard}.
The upper part illustrates the inference-time agentic verification process, where FactGuard assesses uncertainty based on the ambiguity of the input and selectively invokes external tools to acquire additional evidence before refining its reasoning and producing a final decision.
The lower part depicts the training pipeline, which combines supervised fine-tuning with decision-aware reinforcement learning to reinforce structured reasoning, calibrated tool usage, and risk-sensitive verification behavior.
}
\label{fig:pipeline}
\end{figure*}
Early studies on multimodal misinformation detection mainly focused on the image–text setting, where static visual content can be directly aligned with textual claims. These approaches~\cite{eann, qian2021hierarchical, wang2023cross} model cross-modal interactions via attention, feature fusion, or contrastive learning to capture semantic consistency between images and text, achieving promising performance on social media benchmarks.

In recent years, research on misinformation detection has moved beyond static images to the video domain, driven by the growing prevalence of video-based misinformation on social media platforms. Existing approaches~\cite{Qi_2021, fmnv} typically focus on exploiting the rich multimodal signals embedded in videos by jointly modeling visual, acoustic, and textual information. Some studies~\cite{mittal2020emotions, xu2025ample} enhance detection performance by integrating linguistic patterns with emotional or prosodic cues from audio, while others~\cite{wang2025consistency, mccrae2022multi} explicitly model cross-modal inconsistencies among video frames, audio streams, and subtitles. In addition, prior work~\cite{qureshi2021detecting} has explored cross-channel watermarking for manipulation detection, as well as multimodal fusion frameworks that combine topic representations with keyframe-level visual features. More recent efforts~\cite{qi-etal-2023-two, gong2025miningsocialfabricunveiling, fakesv} further incorporate social context or neighborhood structures to capture relational dependencies among multimodal samples.



\subsection{Multimodal LLMs Reasoning}
Recent advances in multimodal large models (MLMs) have significantly improved the ability to jointly understand and reason over visual, textual, and auditory inputs. Models such as GPT-4V~\cite{gpt4v}, LLaVA~\cite{Llava-uhd,Llava-med,Llava-mini,Video-llava}, and Qwen-VL~\cite{qwen,qwen2,qwen3} demonstrate strong generalization across diverse multimodal tasks, enabling more expressive reasoning beyond traditional feature-level fusion.

Building on these models, recent work has increasingly explored multimodal reasoning, including visual grounding~\cite{flamingo}, reasoning process prompting~\cite{chain}, and explanation generation~\cite{unsupervised}. By explicitly modeling intermediate reasoning steps, these methods improve both interpretability and performance on complex perceptual tasks. Reinforcement learning–based post-training has further enhanced model capabilities, as demonstrated by OpenAI-o1~\cite{openai-o1} and DeepSeek-R1~\cite{Deepseek-r1}. 

In misinformation detection, early efforts such as Fact-R1~\cite{fact-r1} demonstrate the promise of combining large pretrained models with domain-specific fine-tuning and reinforcement learning.

\subsection{Tool-Augmented Agentic System} 
Recent works~\cite{zheng2025deepeyes, cui2025t} have increasingly explored augmenting multimodal models with external tools to support more complex reasoning and adaptive information access. Early studies~\cite{mvot, fast} introduce tools as auxiliary sources of visual evidence, enabling models to ground intermediate reasoning steps in external perception signals. Subsequent efforts~\cite{Llava-plus, pyvision} investigate how tool usage can be learned rather than manually specified, through supervision or reinforcement signals that align tool invocation with task objectives. However, tool-augmented and agentic reasoning paradigms remain largely unexplored in video misinformation detection, which is still dominated by single-pass verification without iterative evidence acquisition or adaptive decision-making. In contrast, our work introduces an agentic, tool-augmented verification framework that enables iterative and evidence-driven reasoning.
\section{Methodology}
\label{sec:met}
\subsection{Overview}
As illustrated in Figure~\ref{fig:pipeline}, FactGuard formulates video misinformation verification as an agentic, iterative decision-making process that explicitly integrates multimodal reasoning, evidence-guided action, and outcome-aware optimization. 
In the following sections, we first introduce the agentic formulation of FactGuard and its two-stage inference process (Section~\ref{sub:facgguard}), and then describe the evidence-guided action module that governs tool invocation and information acquisition, including \textit{FactProbe} for external knowledge retrieval and \textit{ClipScout} for targeted video-clip inspection (Section~\ref{sub:tool}). We subsequently detail the training pipeline, including agentic Chain-of-Thought supervised fine-tuning (Section~\ref{sub:sft}), and decision-aware reinforcement learning with structured rewards (Section~\ref{sub:rl}).
\subsection{Problem Formulation}
\label{sub:facgguard}
Each input sample is represented as a triplet
$n=(n^{\mathrm{vid}}, n^{\mathrm{aud}}, n^{\mathrm{txt}})$,
where $n^{\mathrm{vid}}$ denotes the news video content,
$n^{\mathrm{aud}}$ is the speech-to-text transcript extracted from the audio stream,
and $n^{\mathrm{txt}}$ refers to textual metadata such as titles and keywords.

Upon receiving the full multimodal input $n$, the agent applies an agent policy $\mathcal{A}_\theta$ to perform an initial chain-of-thought reasoning pass for assessing the difficulty and ambiguity of the case. The first-stage inference process is abstracted as:
\begin{equation}
    (r_t,\; a_t)=\mathcal{A}_\theta^{(1)}\!\big(n,\, s_t\big),
\end{equation}
where $r_t$ denotes the reasoning trajectory generated at step $t$, and $a_t$ is the agent’s action decision indicating whether to invoke an external tool. The decision is conditioned on the initial observation $n$ together with the agent state $s_t$, which captures its current belief and uncertainty.

If a tool is executed and returns an observation $o_t$, the
agent incorporates this evidence and proceeds to a second-stage, evidence-augmented
reasoning process. This refinement stage jointly considers both the original multimodal input and the tool feedback, and is formalized as:
\begin{equation}
(r_{t+1},\; \hat{y})
=
\mathcal{A}_\theta^{(2)}\!\big(n,\, o_t,\, s_{t+1}\big),
\end{equation}
where $r_{t+1}$ denotes the refined reasoning trajectory and $\hat{y}$ is the final verdict.
The second-stage reasoning conditions on both the original input $n$ and the
tool-provided evidence $o_t$, enabling the agent to update its decision based on
external verification signals. This agentic formulation promotes cautious, evidence-driven verification behaviour, which is
essential for robust misinformation detection under ambiguous and information-scarce
real-world conditions.

\subsection{Evidence-Guided Action Module}
\label{sub:tool}
When the first-stage reasoning identifies missing context or unresolved uncertainty,
FactGuard issues an evidence-guided action to acquire supplementary information. Rather than relying on tools by default, tool invocation is treated
as a deliberate verification decision that is triggered only when internal reasoning
is deemed insufficient. In practice, FactGuard supports multiple evidence acquisition
tools, including \textbf{FactProbe} and \textbf{ClipScout}.

Formally, given an action decision at step $t$, the agent invokes a tool operator
$\mathcal{T}$, which executes the corresponding query and returns an observation:
\begin{equation}
o_t = \mathcal{T}(\kappa_t, \psi_t),
\end{equation}
where $\kappa_t$ specifies the selected tool identity and $\psi_t$ denotes the associated
query parameters.

\paragraph{External Knowledge Retrieval (FactProbe).}
Certain misinformation claims cannot be resolved solely from visual or audio cues, particularly those involving real-world events, historical facts, or temporal assertions. To address this limitation, FactGuard employs \textbf{FactProbe}, a retrieval-augmented knowledge access module that formulates structured factual queries based on the model’s current reasoning state and retrieves evidence from web-based sources. Retrieved results are filtered using reliability heuristics and summarized into a concise textual report, which is incorporated as auxiliary evidence in the refinement stage to supplement incomplete or uncertain textual information and support more reliable verification.

\paragraph{Video-Clip Temporal Inspection (ClipScout).}
Misinformation videos often exhibit long durations with substantial visual redundancy, while only a small number of localized temporal segments contain decisive evidence. ClipScout is designed to enable targeted inspection of such evidence-bearing intervals by selectively sampling representative frames from queried time spans and aggregating them into a compact visual summary. This focused visual evidence supports evidence-oriented reasoning by guiding the model’s attention toward salient events. Through repeated agentic interaction and training, the model learns to efficiently localize and ground critical visual cues without exhaustively processing the entire video.

\subsection{Agentic Supervised Fine-Tuning}
\label{sub:sft}
To address the limited reasoning horizon and underdeveloped tool-usage behaviors of existing multimodal models in misinformation detection, we construct a misinformation-oriented agentic Chain-of-Thought (CoT) dataset that provides structured demonstrations of multi-round reasoning, tool selection intent, and evidence-grounded reflection.

Each annotated trajectory records how the agent analyzes multimodal content, determines when external evidence is required, and incorporates retrieved evidence to refine subsequent reasoning and the final judgment. This reframes misinformation detection from a one-shot classification task into an agentic, self-regulated decision-making process.

We then perform chain-of-thought supervised fine-tuning to align the model with these agentic reasoning behaviors. Given an input $n$ and its annotated agent trajectory $\tau = (r_t, a_t, o_t, r_{t+1}, \hat{y})$, the model is trained to maximize the likelihood of the full reasoning and action sequence:
\begin{equation}
\mathcal{L}_{\mathrm{SFT}}
=
-\mathbb{E}_{(n,\tau)\sim\mathcal{D}}
\big[\log p_\theta(\tau \mid n)\big].
\end{equation}
This supervision encourages the model to internalize longer reasoning chains, explicit uncertainty awareness, and evidence-seeking preferences in ambiguous cases. As a result, FactGuard avoids premature or over-confident conclusions and develops verification-oriented reasoning behaviors that are critical for real-world misinformation detection.

\subsection{Decision-Aware Reinforcement Learning}
\label{sub:rl}
To further calibrate the verification policy under uncertainty, we introduce a decision-aware reinforcement learning stage.
\paragraph{Group Relative Policy Optimization (GRPO).}
We adopt GRPO as a critic-free reinforcement learning paradigm that operates at the level of full agent trajectories. For each input instance, the policy generates a group of candidate trajectories, each reflecting different reasoning depth, tool-invocation behaviour, and final prediction outcomes.
GRPO performs groupwise comparison over these trajectories and updates the policy according to their relative verification quality. The policy network $\pi_\theta$ is optimized by an importance-weighted objective with KL regularization against a frozen reference model:
\begin{equation}
\begin{split}
\mathcal{L}_{GRPO}(\theta) = \mathbb{E}_{q \sim P(Q), \{o_i\}^G_{i=1} \sim \pi_{\theta_{\text{old}}}{(O|q)}}\\
\left[ \sum_{i=1}^G \frac{\pi_\theta(o_i|q)}{\pi_{\theta_{\text{old}}}(o_i|q)} \cdot A_i - \beta \mathbb{D}_{KL}(\pi_\theta||\pi_{ref}) \right],
\end{split}
\end{equation}
\begin{equation}
\mathbb{D}_{KL}(\pi_\theta||\pi_{ref}) = {\frac {\pi_{ref}(o|q)} {{\pi_\theta}(o|q)}} - \log{\frac {\pi_{ref}(o|q)} {{\pi_\theta}(o|q)}} - 1, 
\end{equation}
where $\beta$ controls the trade-off between exploration and stability,
$\pi_{\theta_{\text{old}}}$ denotes the rollout policy used for trajectory sampling, and $\pi_{\mathrm{ref}}$ is a frozen reference model that stabilizes policy updates. The KL term serves as a pointwise surrogate of the KL divergence for efficient policy regularization. Each trajectory $\tau_i$ is assigned a task-specific reward based on verification reliability, including the correctness of the decision and the quality of its reasoning and tool usage. To enable stable comparison within each trajectory group, we compute a normalized advantage:
\begin{equation}
A_i
=
\frac{R(\tau_i)\;-\;\mathrm{mean}(\{R(\tau_j)\})}
     {\mathrm{std}(\{R(\tau_j)\})},
\end{equation}
which reduces reward-scale variance and emphasizes trajectories exhibiting stronger evidence-grounded agent behaviours relative to their peers.
\begin{figure*}[t]
  \centering
  \includegraphics[width=1\linewidth]{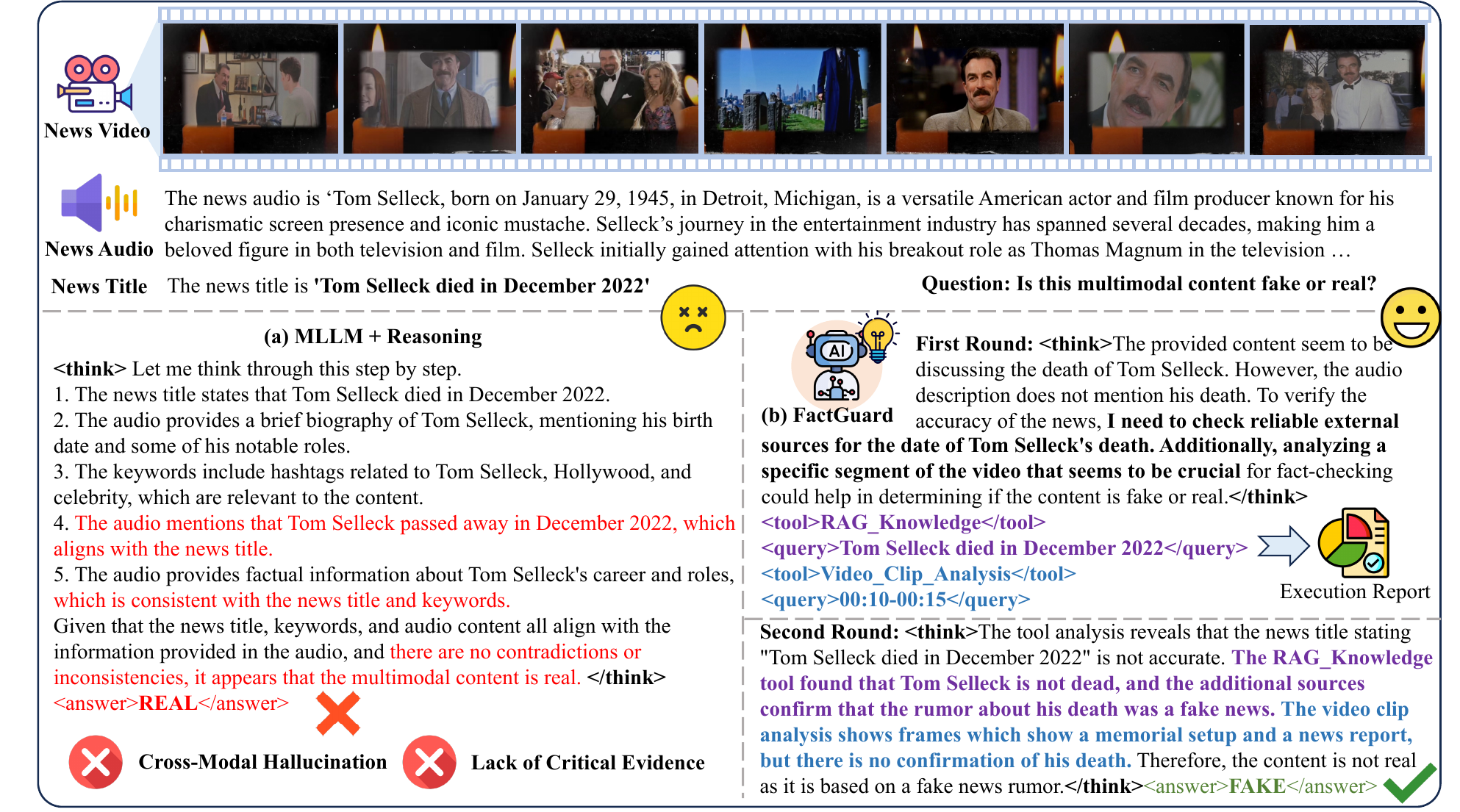}
\caption{\textbf{Key advantages of FactGuard.}
(a) MLLM-based methods with enhanced reasoning may induce cross-modal hallucination in ambiguous cases by over-relying on internally generated assumptions, treating them as grounded evidence without acquiring or validating critical supporting information.
(b) FactGuard formulates misinformation verification as an uncertainty-aware, tool-assisted decision-making process that adaptively refines its conclusions, enabling reliable verification in open and dynamic environments.}
\label{fig:case_2}
\end{figure*}
\paragraph{Reward Design.}
To ensure that reinforcement learning optimizes verification-oriented decision behavior rather than merely maximizing label accuracy, we design a gated reward function that jointly models decision correctness, tool usage, and risk sensitivity for multimodal misinformation detection. We first define the overall trajectory reward and then describe each component in turn. Specifically, for each agent trajectory $\tau$, the total reward is formulated as: 
\begin{equation}
R(\tau)
=
R_{\mathrm{acc}}(\tau)
+
R_{\mathrm{format}}(\tau)
+
\lambda\,R_{\mathrm{risk}}(\tau)
+
R_{\mathrm{tool}}(\tau).
\end{equation}

$R_{\mathrm{acc}}(\tau)$ provides outcome-level supervision based on the final prediction
$\hat{y}$, encouraging correct verification decisions.
To reflect the asymmetric risk profile inherent in misinformation detection, we introduce an explicit risk-shaping term:
\begin{equation}
R_{\mathrm{risk}}(\tau)
=
-\alpha\,\mathbb{I}_{\mathrm{FP}}(\tau)
-\gamma\,\mathbb{I}_{\mathrm{FN}}(\tau),
\end{equation}
where $\mathbb{I}_{\mathrm{FP}}(\tau)$ and $\mathbb{I}_{\mathrm{FN}}(\tau)$ indicate false-positive and false-negative outcomes, respectively. The coefficients $\alpha$ and $\gamma$ explicitly control the trade-off between precision and recall, allowing the detector’s risk preference to be adjusted for different misinformation scenarios.

Beyond decision outcomes, reinforcement learning is constrained to valid and interpretable agent behaviors. Accordingly, $R_{\mathrm{format}}(\tau)$ enforces well-formed outputs by requiring structured reasoning and final decisions to be enclosed within \textless think\textgreater{} and \textless answer\textgreater{} tags, thereby restricting policy optimization to well-defined agent trajectories.

In addition to supervising decision correctness and output validity, we introduce an action-level reward to explicitly regulate tool usage. Tool invocation is treated as a deliberate verification action and is rewarded only when it leads to a correct verification outcome, while unnecessary or ineffective tool use is penalized:
\begin{equation}
R_{\mathrm{tool}}(\tau)
=
\begin{cases}
\;\;\;r_{\mathrm{tool}}^{+}, & \text{tool used and } R_{\mathrm{acc}}(\tau)>0,\\
-\,r_{\mathrm{tool}}^{-}, & \text{tool used and } R_{\mathrm{acc}}(\tau)\le 0,\\
0, & \text{otherwise}.
\end{cases}
\end{equation}
This design discourages indiscriminate tool usage while rewarding evidence-seeking behavior only when it contributes to correct verification. Consequently, GRPO optimizes verification-oriented agent behavior under uncertainty beyond raw label accuracy, including calibrated tool invocation, evidence-grounded reasoning, and adaptive risk control. Unlike prior reinforcement learning methods that focus solely on output correctness, our approach treats misinformation detection as a decision-aware agentic problem, in which uncertainty handling and tool usage are first-class optimization objectives.

\begin{table*}[t]
\centering\small
\caption{Performance comparison on FakeSV, FakeTT and FakeVV datasets. We highlight the \colorbox{lightpurple}{improvements} achieved by FactGuard.}
\resizebox{\textwidth}{!}{
\begin{tabular}{l|cccc|cccc|cccc}
\toprule
\multirow{2}{*}{\textbf{Model}} & \multicolumn{4}{c|}{\textbf{FakeSV}} & \multicolumn{4}{c|}{\textbf{FakeTT}} & \multicolumn{4}{c}{\textbf{FakeVV}} \\
 & Acc & Prec & Rec & F1 & Acc & Prec & Rec & F1 & Acc & Prec & Rec & F1 \\
\midrule
BERT \cite{bert} & 65.4 & 66.0 & 66.5 & 66.2 & 68.7 & 67.5 & 67.5 & 67.5 & 60.4 & 57.9 & 56.8 & 57.3 \\
\midrule
TikTec \cite{tiktec} & 64.8 & 63.2 & 61.9 & 62.5 & 61.1 & 64.8 & 64.2 & 64.5 & 59.3 & 59.1 & 59.5 & 59.3 \\
FANVM \cite{fanvm} & 65.4 & 66.1 & 64.3 & 65.2 & 68.9 & 64.7 & 68.8 & 67.1 & 61.9 & 60.7 & 60.8 & 60.8 \\
SV-FEND \cite{fakesv} & 67.1 & 67.4 & 66.3 & 66.8 & 67.6 & 72.2 & \underline{69.0} & 70.6 & 70.9 & 71.4 & 71.3 & 71.3 \\
FakingRec \cite{fakett} & 69.5 & 69.7 & 70.4 & 70.0 & 71.0 & 71.9 & 72.0 & 72.0 & 72.1 & 72.4 & 71.6 & 72.0 \\
\midrule
Gemini2-thinking \cite{gemini2thinking} & 63.1 & 61.8 & 61.9 & 61.9 & 56.6 & 55.2 & 55.3 & 55.3 & 51.5 & 46.0 & 46.0 & 48.6 \\
GPT-4o \cite{gpt4o} & 66.6 & 65.2 & 64.7 & 64.9 & 57.9 & 57.8 & 62.9 & 60.2 & 56.0 & 60.4 & 35.0 & 44.3 \\
GPT-o1-mini \cite{gpt4omini} & 60.3 & 57.7 & 56.5 & 57.1 & 52.5 & 51.6 & 51.7 & 51.7 & 47.5 & 46.9 & 37.6 & 41.8 \\
\midrule
DeepSeek-R1 \cite{Deepseek-r1} & 61.8 & 60.4 & 60.3 & 60.3 & 49.8 & 52.6 & 52.5 & 52.6 & 53.5 & 58.1 & 25.2 & 35.1 \\
Qwen2.5-VL-7B \cite{qwen2.5} & 55.6 & 55.5 & 55.7 & 55.6 & 54.9 & 54.0 & 54.1 & 54.0 & 52.9 & 51.1 & 51.1 & 51.1 \\
Qwen2.5-VL-72B \cite{qwen2.5} & 57.6 & 55.4 & 55.2 & 55.3 & 59.2 & 58.1 & 58.3 & 58.2 & 54.0 & 60.0 & 24.0 & 34.3 \\
QVQ-72B-preview \cite{qvq} & 60.8 & 59.0 & 58.8 & 58.9 & 58.1 & 54.0 & 52.8 & 53.4 & 53.5 & 52.6 & 52.6 & 52.6 \\
InternVL2.5-8B \cite{Internvl2.5} & 49.8 & 52.6 & 52.5 & 52.6 & 43.9 & 44.0 & 44.0 & 44.0 & 53.5 & 58.5 & 24.0 & 34.0 \\
InternVL2.5-78B-MPO \cite{intervl-mpo} & 57.5 & 53.0 & 52.0 & 52.5 & 59.2 & 57.1 & 56.7 & 56.9 & 54.0 & 60.0 & 24.0 & 34.3 \\
\midrule
Fact-R1~\cite{fact-r1}& \underline{75.6} & \underline{77.7} & \underline{72.0} & \underline{74.7} & \underline{74.4} & {\textbf{77.8}} & 68.3 & \underline{72.7} & \underline{81.2} & \underline{84.5} & \underline{76.4} & \underline{80.3} \\

\rowcolor{lightpurple}
\textbf{FactGuard (Ours)} & {\textbf{79.3}} & {\textbf{82.2}} & {\textbf{80.6}} & {\textbf{81.4}} & {\textbf{75.3}} & \underline{73.8} & {\textbf{76.7}} & {\textbf{75.2}} & {\textbf{83.0}} & \textbf{85.8} & {\textbf{82.1}} & {\textbf{83.9}}\\

\bottomrule
\end{tabular}
}
\label{tab:comparation}
\end{table*}
\section{Experiment}
\label{sec:exp}

\subsection{Experimental Settings}
\paragraph{Baselines.} To comprehensively evaluate the performance of FactGuard, we compare it against three categories of baselines. \textbf{Discriminative Models} include the single-modality method, BERT~\cite{bert} and multimodal methods such as TikTec~\cite{tiktec}, FANVM~\cite{fanvm}, SVFEND~\cite{fakesv}, and FakingRec~\cite{fakett}. \textbf{Zero-shot MLLMs} comprise closed-source models, including Gemini2-thinking~\cite{gemini2thinking}, GPT-4o~\cite{gpt4o}, and GPT-o1-mini~\cite{gpt4omini}, as well as open-source models, including Qwen2.5-VL~\cite{qwen2.5}, InternVL2.5~\cite{Internvl2.5}, QVQ-72B~\cite{qvq}, InternVL2.5-MPO~\cite{intervl-mpo}, and DeepSeek-R1~\cite{Deepseek-r1}, which are evaluated in a zero-shot setting. For models without native video support, textual descriptions of news videos are used as substitutes. Finally, \textbf{Task-Aligned Reasoning Models} include Fact-R1~\cite{fact-r1} and our method.
\paragraph{Benchmarks.} We employ three widely adopted benchmark datasets: FakeSV~\cite{fakesv}, FakeTT~\cite{fakett}, and FakeVV~\cite{fact-r1}. Following the protocol in~\cite{fakesv}, we use a temporal split for testing, selecting the most recent 15\% of samples from each dataset. In accordance with the Fact-R1 standard, all baseline models are trained on the training sets of FakeVV, FakeTT, and FakeSV. We report four standard evaluation metrics, including accuracy (ACC), precision, recall, and F1 score, to provide a comprehensive assessment of classification performance.

\paragraph{Training Details.} Our model is trained on Qwen2.5-VL-7B using a two-stage pipeline consisting of supervised fine-tuning (SFT) followed by decision-aware reinforcement learning with GRPO. All experiments are conducted on 8 NVIDIA H100 GPUs. During GRPO training, each input prompt is unfolded into 8 candidate trajectories. The video-clip inspection tool can be invoked at most once per prompt, while external knowledge retrieval is unconstrained. We set the learning rate to $1\times10^{-6}$ and train the policy for one epoch. The maximum prompt length is set to 16,384 tokens, and the maximum response length is 768 tokens. The KL regularization coefficient is set to 0.04. Unless otherwise specified, we use a per-device batch size of 1 with no gradient accumulation. Additional implementation details are provided in Appendix~\ref{sub:exdetail}.

\subsection{Main Results}
\paragraph{Comparison to State-of-the-Art Approaches.}
As shown in Table~\ref{tab:comparation}, we evaluate FactGuard on the FakeSV, FakeTT, and FakeVV datasets and compare it with discriminative models, zero-shot MLLMs, and task-aligned reasoning models specifically trained or reinforced for misinformation detection. All results are reported using accuracy, precision, recall, and F1-score. Overall, FactGuard consistently outperforms existing methods across all datasets in terms of both accuracy and F1 score.

For discriminative approaches, FakingRec yields the strongest performance, benefiting from its effective modeling of fine-grained multimodal correlations and its exploitation of user--content interaction patterns, which are particularly suited for recommendation-style verification. For zero-shot MLLMs, GPT-4o attains the highest accuracy, likely due to its large-scale multimodal pretraining and strong general reasoning priors. Within the same model family, larger models generally exhibit better performance, although their effectiveness varies across datasets, reflecting differences in task difficulty and evidence availability.

Compared to discriminative methods, task-aligned reasoning approaches, including Fact-R1 and our method, not only provide improved interpretability through explicit reasoning processes but also achieve substantially higher verification accuracy. While Fact-R1 significantly improves performance via reinforcement learning, FactGuard consistently delivers superior results. As illustrated in Figure~\ref{fig:case_2}(b), this advantage stems from FactGuard’s ability to avoid cross-modal hallucination in ambiguous cases by explicitly recognizing uncertainty and adaptively acquiring supporting evidence through tool-assisted, iterative decision making.

\begin{figure}[t]
  \centering
  \includegraphics[width=1\linewidth]{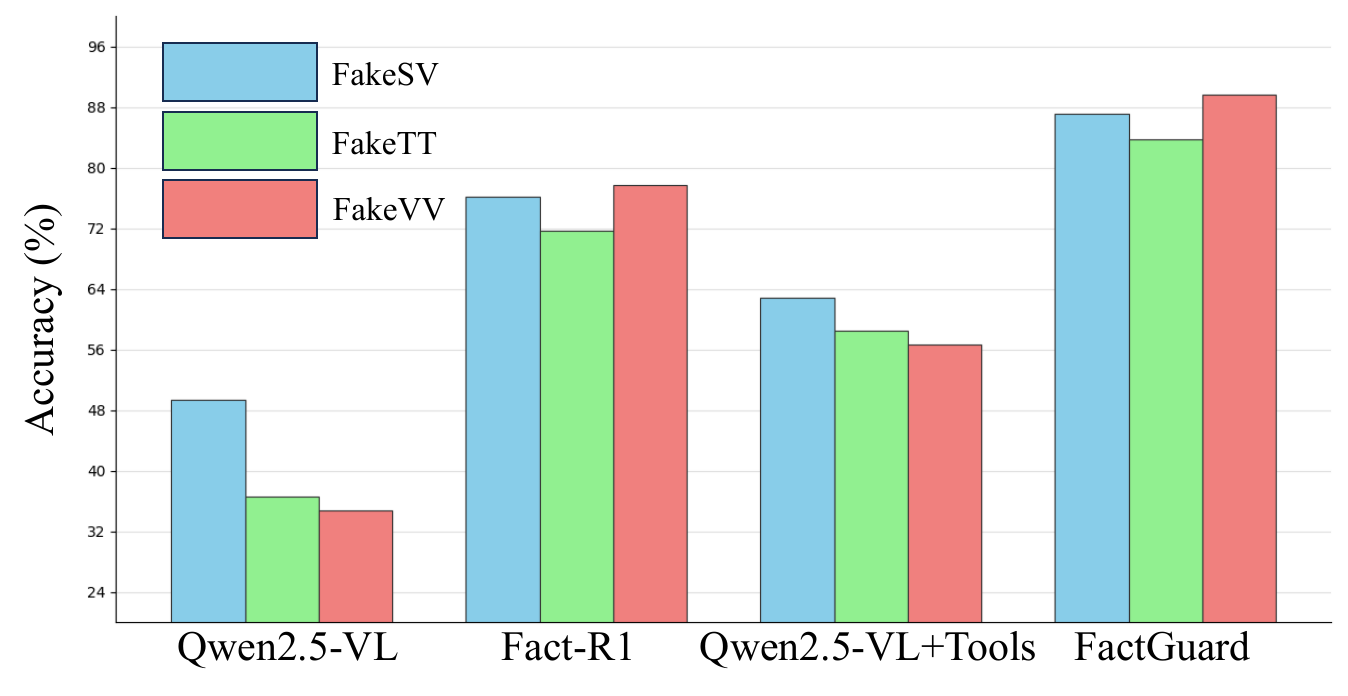}
\caption{
\textbf{Qualitative analysis of model reasoning.}
Representative reasoning traces under correct predictions show that FactGuard produces more coherent and evidence-grounded reasoning than Qwen2.5-VL and Fact-R1, highlighting improved interpretability.
}
\label{fig:explain}
\end{figure}
\paragraph{Explainability Analysis.} Compared to discriminative models, multimodal large language model–based approaches not only produce final predictions but also generate explicit reasoning traces, substantially improving interpretability. Accordingly, we adopt \textbf{GPT-4o} as an automatic evaluator to assess the quality of model-generated reasoning and to provide fine-grained reasoning accuracy scores. The evaluation considers multiple complementary dimensions, including faithfulness to the provided multimodal evidence, logical consistency of the reasoning chain, and the ability to capture salient misinformation patterns. 

To ensure a fair and meaningful evaluation, we restrict the analysis to instances where the predicted labels are correct, thereby preventing incorrect predictions from confounding the assessment of reasoning quality. As illustrated in Figure~\ref{fig:explain}, we compare Qwen2.5-VL and Fact-R1 without tool assistance against their tool-augmented counterparts, Qwen2.5-VL with tools and FactGuard. The results indicate that both reinforcement learning and external tool integration contribute to improved reasoning capability over the base model. Reinforcement learning enhances the model’s ability to generate structured and consistent reasoning, while tool integration further strengthens evidence grounding. Notably, their combination yields a complementary effect: FactGuard consistently exhibits the strongest reasoning performance, producing coherent, evidence-grounded, and logically sound inference chains across diverse cases.

\begin{table}[t]
\centering\small
\caption{Cost-sensitive precision and recall achieved by FactGuard on FakeSV and FakeVV under different asymmetric error cost ratios.}
\begin{tabular}{c|cc|cc}
\toprule
\multicolumn{1}{c|}{\textbf{Cost Ratio}} 
& \multicolumn{2}{c|}{\textbf{FakeSV}} 
& \multicolumn{2}{c}{\textbf{FakeVV}} \\
\multicolumn{1}{c|}{\textbf{($\alpha:\gamma$)}}  & Precision & Recall & Precision & Recall \\
\midrule
1:2 & 80.8 & \textbf{82.1} & 83.2 & \textbf{83.7} \\
1:1 & 82.2 & 80.6 & 85.8 & 82.1 \\
2:1 & \textbf{83.2} & 75.0 & \textbf{86.6} & 79.3 \\
\bottomrule
\end{tabular}
\label{tab:cost_sensitive_pr_rec}
\end{table}

\paragraph{Cost-Sensitive Risk Analysis.}
We report the precision and recall on the FakeSV and FakeTT datasets to analyze the effect of cost-sensitive risk modeling. As shown in the Table~\ref{tab:cost_sensitive_pr_rec}, the cost ratio $\alpha:\gamma$ plays a critical role in shaping the trade-off between precision and recall. When $\alpha$ is increased, the model places a higher penalty on false positives, leading to improved \textit{Fake} precision at the cost of a modest reduction in recall. Conversely, emphasizing $\gamma$ encourages the model to reduce false negatives, resulting in higher recall but lower precision. This behavior highlights the flexibility of the proposed cost-sensitive risk function in regulating decision preferences under asymmetric error costs. Unless otherwise specified, we adopt a balanced cost ratio of $\alpha:\gamma = 1:1$ in all other experiments.

Unlike conventional misinformation detectors that optimize a fixed objective, FactGuard enables explicit control over the precision–recall balance, allowing the verification policy to be adapted to different deployment requirements. 
In practice, this is particularly important for misinformation detection, where the relative costs of false alarms and missed misinformation can vary substantially across application scenarios. For example, high-precision settings are desirable in content moderation to avoid unjustified censorship, whereas high-recall configurations are preferable in early-warning or monitoring systems to minimize the spread of harmful content. These results demonstrate that FactGuard offers a principled and practical approach to risk-aware verification, effectively bridging model optimization with real-world decision-making constraints.
\begin{table}[t]
\centering\small
\caption{Ablation study of FactGuard on FakeSV and FakeTT datasets. We report Accuracy (\%) and F1-score (\%) to evaluate the contribution of each component.}
\label{tab:ablation_study}
\resizebox{0.87\linewidth}{!}{
\begin{tabular}{l|cc|cc}
\toprule
\multirow{2}{*}{\textbf{Model}} & \multicolumn{2}{c|}{\textbf{FakeSV}} & \multicolumn{2}{c}{\textbf{FakeTT}} \\
 & Acc & F1 & Acc & F1 \\
\midrule
\addlinespace[-0.001ex]
\rowcolor{lightpurple}
\textbf{FactGuard (Ours)} & \textbf{79.3} & \textbf{81.4} & \textbf{75.3} & \textbf{75.2} \\
\quad w/o SFT & 73.1 & 74.7 & 71.6 & 68.2 \\
\quad w/o RL & 62.0 & 63.2 & 67.9 & 65.7 \\
\quad w/o $R_{tool}$  & 77.7 & 78.5 & 74.7 & 72.9 \\
\quad w/o $R_{risk}$  & 78.5 & 78.9 & 74.9 & 73.8 \\
\midrule
\textit{Base (Qwen2.5-VL-7B+tool)} & 57.6 & 60.6 & 56.8 & 55.4 \\
\bottomrule
\end{tabular}
}
\end{table}
\subsection{Ablation Studies}
\paragraph{Variants of component ablations.}
Table~\ref{tab:ablation_study} presents the ablation results of FactGuard on the FakeSV and FakeTT datasets. The full model consistently achieves the best performance, indicating that all components contribute positively to misinformation detection. Removing either supervised fine-tuning (SFT) or reinforcement learning (RL) leads to substantial performance degradation, with the absence of RL causing the most severe drop, highlighting its critical role in optimizing multi-step verification and decision-making. In contrast, ablating the tool-related reward ($R_{\mathrm{tool}}$) or the risk-aware reward ($R_{\mathrm{risk}}$) results in moderate but consistent performance declines, suggesting that these rewards provide complementary benefits by improving evidence grounding and uncertainty calibration. 

We find that reinforcement learning without prior SFT yields inferior performance, consistent with Video-Star~\cite{video-star}. Direct RL causes the policy to collapse toward tool-free reasoning, as it must simultaneously explore complex reasoning trajectories and tool-use decisions under distributional mismatch. This motivates our two-stage training scheme, where SFT establishes structured evidence-seeking behaviors before RL refinement. 

\paragraph{Variants of different GRPO hyperparameters.}As shown in Table~\ref{tab:ablation_grpo_fakesv}, increasing the number of rollouts generally improves performance by yielding more reliable advantage estimates, with diminishing returns beyond 8 rollouts, while further increasing the number of rollouts incurs additional computational overhead with limited efficiency gains. We also observe that the KL coefficient $\beta$ exhibits a clear trade-off: overly small values lead to insufficient regularization, while larger values overly constrain policy updates. Based on this balance, we adopt 8 rollouts and $\beta$ = 0.04.

\begin{table}[t]
\centering\small
\caption{Performance of GRPO variants on FakeSV and FakeTT.}
\label{tab:ablation_grpo_fakesv}
\resizebox{0.98\linewidth}{!}{
\begin{tabular}{l|cc|cc}
\toprule
\multirow{2}{*}{\textbf{Variant}} 
& \multicolumn{2}{c|}{\textbf{FakeSV}} 
& \multicolumn{2}{c}{\textbf{FakeTT}} \\
 & Acc & F1 & Acc & F1 \\
\midrule
\emph{(a)} w/. Rollouts = 6
& 77.3 & 78.9 
& 72.6 & 69.4 \\
\rowcolor{lightpurple}
\emph{(b)} w/. Rollouts = 8 (Ours)  
& 79.3 & 81.4 
& \textbf{75.3} & \textbf{75.2} \\
\emph{(c)} w/. Rollouts = 12
& \textbf{79.5} & \textbf{81.5} 
& 75.2 & 74.3 \\
\midrule
\emph{(d)} w/. $\beta$ = 0.02 
& 78.2 & 79.5 
& 74.6 & 73.4 \\
\rowcolor{lightpurple}
\emph{(e)} w/. $\beta$ = 0.04 (Ours) 
& \textbf{79.3} & \textbf{81.4} 
& \textbf{75.3} & \textbf{75.2} \\
\emph{(f)} w/. $\beta$ = 0.06 
& 78.8 & 80.9 
& 74.8 & 73.7 \\
\bottomrule
\end{tabular}
}
\end{table}

\section{Conclusion}
\label{sec:con}
In this work, we presented \textbf{FactGuard}, an agentic framework for video misinformation detection that formulates verification as an uncertainty-aware, iterative decision-making process. By integrating multimodal reasoning with selective, tool-assisted evidence acquisition, FactGuard addressed key limitations of existing approaches, including over-reliance on internally generated assumptions and the failure to obtain critical supporting evidence in ambiguous cases. We further introduced a two-stage training strategy that combines agentic Chain-of-Thought supervised fine-tuning with decision-aware reinforcement learning, enabling high-quality reasoning, calibrated tool usage, and explicit control over asymmetric verification risks. Extensive experiments on three public benchmarks demonstrate that FactGuard consistently outperforms strong discriminative baselines and MLLM-based reasoning models, including reinforcement learning–enhanced methods, in both prediction accuracy and reasoning quality.

\section*{Impact Statement}
This work aims to advance research in multimodal reasoning, agentic decision-making, and video misinformation detection. The proposed framework is developed solely for research purposes and is intended to support the study of reliable and interpretable verification systems. All datasets used in this work are publicly available and are employed in accordance with their respective licenses and intended usage scopes. Any additional data constructed for training or evaluation are synthetically generated or obtained through automated processes, and are used exclusively for research purposes. While the techniques studied in this work could potentially be applied in real-world verification or content moderation scenarios, we emphasize that our contributions are intended as research tools rather than deployment-ready systems. We do not foresee significant negative societal impact when the proposed methods are used responsibly and within their intended research scope.

\bibliography{example_paper}

\begin{thebibliography}{49}
\providecommand{\natexlab}[1]{#1}
\providecommand{\url}[1]{\texttt{#1}}
\expandafter\ifx\csname urlstyle\endcsname\relax
  \providecommand{\doi}[1]{doi: #1}\else
  \providecommand{\doi}{doi: \begingroup \urlstyle{rm}\Url}\fi

\bibitem[Achiam et~al.(2023)Achiam, Adler, Agarwal, Ahmad, Akkaya, Aleman, Almeida, Altenschmidt, Altman, Anadkat, et~al.]{gpt4o}
Achiam, J., Adler, S., Agarwal, S., Ahmad, L., Akkaya, I., Aleman, F.~L., Almeida, D., Altenschmidt, J., Altman, S., Anadkat, S., et~al.
\newblock Gpt-4 technical report.
\newblock \emph{arXiv preprint arXiv:2303.08774}, 2023.

\bibitem[Alayrac et~al.(2022)Alayrac, Donahue, Luc, Miech, Barr, Hasson, Lenc, Mensch, Millican, Reynolds, et~al.]{flamingo}
Alayrac, J.-B., Donahue, J., Luc, P., Miech, A., Barr, I., Hasson, Y., Lenc, K., Mensch, A., Millican, K., Reynolds, M., et~al.
\newblock Flamingo: a visual language model for few-shot learning.
\newblock \emph{Advances in neural information processing systems}, 35:\penalty0 23716--23736, 2022.

\bibitem[Bai et~al.(2023)Bai, Bai, Chu, Cui, Dang, Deng, Fan, Ge, Han, Huang, Hui, Ji, Li, Lin, Lin, Liu, Liu, Lu, Lu, Ma, Men, Ren, Ren, Tan, Tan, Tu, Wang, Wang, Wang, Wu, Xu, Xu, Yang, Yang, Yang, Yang, Yao, Yu, Yuan, Yuan, Zhang, Zhang, Zhang, Zhang, Zhou, Zhou, Zhou, and Zhu]{qwen}
Bai, J., Bai, S., Chu, Y., Cui, Z., Dang, K., Deng, X., Fan, Y., Ge, W., Han, Y., Huang, F., Hui, B., Ji, L., Li, M., Lin, J., Lin, R., Liu, D., Liu, G., Lu, C., Lu, K., Ma, J., Men, R., Ren, X., Ren, X., Tan, C., Tan, S., Tu, J., Wang, P., Wang, S., Wang, W., Wu, S., Xu, B., Xu, J., Yang, A., Yang, H., Yang, J., Yang, S., Yao, Y., Yu, B., Yuan, H., Yuan, Z., Zhang, J., Zhang, X., Zhang, Y., Zhang, Z., Zhou, C., Zhou, J., Zhou, X., and Zhu, T.
\newblock Qwen technical report, 2023.
\newblock URL \url{https://arxiv.org/abs/2309.16609}.

\bibitem[Bai et~al.(2025)Bai, Chen, Liu, Wang, Ge, Song, Dang, Wang, Wang, Tang, et~al.]{qwen2.5}
Bai, S., Chen, K., Liu, X., Wang, J., Ge, W., Song, S., Dang, K., Wang, P., Wang, S., Tang, J., et~al.
\newblock Qwen2. 5-vl technical report.
\newblock \emph{arXiv preprint arXiv:2502.13923}, 2025.

\bibitem[Bu et~al.(2023)Bu, Sheng, Cao, Qi, Wang, and Li]{bu2023combating}
Bu, Y., Sheng, Q., Cao, J., Qi, P., Wang, D., and Li, J.
\newblock Combating online misinformation videos: Characterization, detection, and future directions.
\newblock In \emph{Proceedings of the 31st ACM International Conference on Multimedia}, pp.\  8770--8780, 2023.

\bibitem[Bu et~al.(2024)Bu, Sheng, Cao, Qi, Wang, and Li]{fakett}
Bu, Y., Sheng, Q., Cao, J., Qi, P., Wang, D., and Li, J.
\newblock Fakingrecipe: Detecting fake news on short video platforms from the perspective of creative process.
\newblock In \emph{Proceedings of the 32nd ACM International Conference on Multimedia}, pp.\  1351–1360. Association for Computing Machinery, 2024.
\newblock \doi{10.1145/3664647.3680663}.

\bibitem[Chen et~al.(2024)Chen, Wu, Wang, Su, Chen, Xing, Zhong, Zhang, Zhu, Lu, et~al.]{Internvl2.5}
Chen, Z., Wu, J., Wang, W., Su, W., Chen, G., Xing, S., Zhong, M., Zhang, Q., Zhu, X., Lu, L., et~al.
\newblock Internvl: Scaling up vision foundation models and aligning for generic visual-linguistic tasks.
\newblock In \emph{Proceedings of the IEEE/CVF conference on computer vision and pattern recognition}, pp.\  24185--24198, 2024.

\bibitem[Cheng et~al.(2023)Cheng, Wu, Chen, Li, Liu, and Kong]{unsupervised}
Cheng, S., Wu, Z., Chen, J., Li, Z., Liu, Y., and Kong, L.
\newblock Unsupervised explanation generation via correct instantiations.
\newblock In \emph{Proceedings of the AAAI conference on artificial intelligence}, volume~37, pp.\  12700--12708, 2023.

\bibitem[Choi \& Ko(2021)Choi and Ko]{fanvm}
Choi, H. and Ko, Y.
\newblock Using topic modeling and adversarial neural networks for fake news video detection.
\newblock In \emph{Proceedings of the 30th ACM international conference on information \& knowledge management}, pp.\  2950--2954, 2021.

\bibitem[Cui et~al.(2025)Cui, Zou, Li, Li, Xu, Liu, and Huang]{cui2025t}
Cui, X., Zou, Y., Li, Z., Li, P., Xu, X., Liu, X., and Huang, H.
\newblock T\^{} 2agent a tool-augmented multimodal misinformation detection agent with monte carlo tree search.
\newblock \emph{arXiv preprint arXiv:2505.19768}, 2025.

\bibitem[{Gemini Team}(2023)]{gemini2thinking}
{Gemini Team}.
\newblock Gemini: a family of highly capable multimodal models.
\newblock \emph{arXiv preprint arXiv:2312.11805}, 2023.

\bibitem[Gong et~al.(2025)Gong, Su, Zhang, Li, Liu, Wu, and Wang]{gong2025miningsocialfabricunveiling}
Gong, H., Su, B., Zhang, X., Li, J., Liu, Q., Wu, S., and Wang, L.
\newblock Mining the social fabric: Unveiling communities for fake news detection in short videos, 2025.
\newblock URL \url{https://arxiv.org/abs/2508.07992}.

\bibitem[Guo et~al.(2025)Guo, Yang, Zhang, Song, Zhang, Xu, Zhu, Ma, Wang, Bi, et~al.]{Deepseek-r1}
Guo, D., Yang, D., Zhang, H., Song, J., Zhang, R., Xu, R., Zhu, Q., Ma, S., Wang, P., Bi, X., et~al.
\newblock Deepseek-r1: Incentivizing reasoning capability in llms via reinforcement learning.
\newblock \emph{arXiv preprint arXiv:2501.12948}, 2025.

\bibitem[Guo et~al.(2024)Guo, Xu, Yao, Cui, Ni, Ge, Chua, Liu, and Huang]{Llava-uhd}
Guo, Z., Xu, R., Yao, Y., Cui, J., Ni, Z., Ge, C., Chua, T.-S., Liu, Z., and Huang, G.
\newblock Llava-uhd: an lmm perceiving any aspect ratio and high-resolution images.
\newblock In \emph{European Conference on Computer Vision}, pp.\  390--406. Springer, 2024.

\bibitem[Hou et~al.(2019)Hou, P{\'e}rez-Rosas, Loeb, and Mihalcea]{hou2019towards}
Hou, R., P{\'e}rez-Rosas, V., Loeb, S., and Mihalcea, R.
\newblock Towards automatic detection of misinformation in online medical videos.
\newblock In \emph{2019 International conference on multimodal interaction}, pp.\  235--243, 2019.

\bibitem[Jaech et~al.(2024{\natexlab{a}})Jaech, Kalai, Lerer, Richardson, El-Kishky, Low, Helyar, Madry, Beutel, Carney, et~al.]{gpt4omini}
Jaech, A., Kalai, A., Lerer, A., Richardson, A., El-Kishky, A., Low, A., Helyar, A., Madry, A., Beutel, A., Carney, A., et~al.
\newblock Openai o1 system card.
\newblock \emph{arXiv preprint arXiv:2412.16720}, 2024{\natexlab{a}}.

\bibitem[Jaech et~al.(2024{\natexlab{b}})Jaech, Kalai, Lerer, Richardson, El-Kishky, Low, Helyar, Madry, Beutel, Carney, et~al.]{openai-o1}
Jaech, A., Kalai, A., Lerer, A., Richardson, A., El-Kishky, A., Low, A., Helyar, A., Madry, A., Beutel, A., Carney, A., et~al.
\newblock Openai o1 system card.
\newblock \emph{arXiv preprint arXiv:2412.16720}, 2024{\natexlab{b}}.

\bibitem[Knuutila et~al.(2021)Knuutila, Herasimenko, Au, Bright, and Howard]{knuutila2021dataset}
Knuutila, A., Herasimenko, A., Au, H., Bright, J., and Howard, P.~N.
\newblock A dataset of covid-related misinformation videos and their spread on social media.
\newblock \emph{Journal of Open Humanities Data}, 7, 2021.

\bibitem[Koroteev(2021)]{bert}
Koroteev, M.~V.
\newblock Bert: a review of applications in natural language processing and understanding.
\newblock \emph{arXiv preprint arXiv:2103.11943}, 2021.

\bibitem[Li et~al.(2023)Li, Wong, Zhang, Usuyama, Liu, Yang, Naumann, Poon, and Gao]{Llava-med}
Li, C., Wong, C., Zhang, S., Usuyama, N., Liu, H., Yang, J., Naumann, T., Poon, H., and Gao, J.
\newblock Llava-med: Training a large language-and-vision assistant for biomedicine in one day.
\newblock \emph{Advances in Neural Information Processing Systems}, 36:\penalty0 28541--28564, 2023.

\bibitem[Li et~al.(2025)Li, Wu, Zhang, Xia, Mao, Dong, Vulić, and Wei]{mvot}
Li, C., Wu, W., Zhang, H., Xia, Y., Mao, S., Dong, L., Vulić, I., and Wei, F.
\newblock Imagine while reasoning in space: Multimodal visualization-of-thought, 2025.
\newblock URL \url{https://arxiv.org/abs/2501.07542}.

\bibitem[Lin et~al.(2024)Lin, Ye, Zhu, Cui, Ning, Jin, and Yuan]{Video-llava}
Lin, B., Ye, Y., Zhu, B., Cui, J., Ning, M., Jin, P., and Yuan, L.
\newblock Video-llava: Learning united visual representation by alignment before projection.
\newblock In \emph{Proceedings of the 2024 conference on empirical methods in natural language processing}, pp.\  5971--5984, 2024.

\bibitem[Liu et~al.(2024)Liu, Cheng, Liu, Zhang, Li, Ren, Zou, Yang, Su, Zhu, Zhang, Gao, and Li]{Llava-plus}
Liu, S., Cheng, H., Liu, H., Zhang, H., Li, F., Ren, T., Zou, X., Yang, J., Su, H., Zhu, J., Zhang, L., Gao, J., and Li, C.
\newblock Llava-plus: Learning to use tools for creating multimodal agents.
\newblock In \emph{European Conference on Computer Vision}, pp.\  126--142. Springer, 2024.

\bibitem[McCrae et~al.(2022)McCrae, Wang, and Zakhor]{mccrae2022multi}
McCrae, S., Wang, K., and Zakhor, A.
\newblock Multi-modal semantic inconsistency detection in social media news posts.
\newblock In \emph{International Conference on Multimedia Modeling}, pp.\  331--343. Springer, 2022.

\bibitem[Mittal et~al.(2020)Mittal, Bhattacharya, Chandra, Bera, and Manocha]{mittal2020emotions}
Mittal, T., Bhattacharya, U., Chandra, R., Bera, A., and Manocha, D.
\newblock Emotions don't lie: An audio-visual deepfake detection method using affective cues.
\newblock In \emph{Proceedings of the 28th ACM international conference on multimedia}, pp.\  2823--2832, 2020.

\bibitem[Qi et~al.(2021)Qi, Cao, Li, Liu, Sheng, Mi, He, Lv, Guo, and Yu]{Qi_2021}
Qi, P., Cao, J., Li, X., Liu, H., Sheng, Q., Mi, X., He, Q., Lv, Y., Guo, C., and Yu, Y.
\newblock Improving fake news detection by using an entity-enhanced framework to fuse diverse multimodal clues.
\newblock In \emph{Proceedings of the 29th ACM International Conference on Multimedia}. ACM, 2021.
\newblock \doi{10.1145/3474085.3481548}.

\bibitem[Qi et~al.(2023{\natexlab{a}})Qi, Bu, Cao, Ji, Shui, Xiao, Wang, and Chua]{fakesv}
Qi, P., Bu, Y., Cao, J., Ji, W., Shui, R., Xiao, J., Wang, D., and Chua, T.-S.
\newblock Fakesv: A multimodal benchmark with rich social context for fake news detection on short video platforms.
\newblock In \emph{Proceedings of the AAAI Conference on Artificial Intelligence}, volume~37, pp.\  14444--14452, 2023{\natexlab{a}}.

\bibitem[Qi et~al.(2023{\natexlab{b}})Qi, Zhao, Shen, Ji, Cao, and Chua]{qi-etal-2023-two}
Qi, P., Zhao, Y., Shen, Y., Ji, W., Cao, J., and Chua, T.-S.
\newblock Two heads are better than one: Improving fake news video detection by correlating with neighbors.
\newblock In \emph{Findings of the Association for Computational Linguistics: ACL 2023}, pp.\  11947--11959. Association for Computational Linguistics, 2023{\natexlab{b}}.
\newblock \doi{10.18653/v1/2023.findings-acl.756}.

\bibitem[Qian et~al.(2021)Qian, Wang, Hu, Fang, and Xu]{qian2021hierarchical}
Qian, S., Wang, J., Hu, J., Fang, Q., and Xu, C.
\newblock Hierarchical multi-modal contextual attention network for fake news detection.
\newblock In \emph{Proceedings of the 44th international ACM SIGIR conference on research and development in information retrieval}, pp.\  153--162, 2021.

\bibitem[Qureshi et~al.(2021)Qureshi, Meg{\'\i}as, and Kuribayashi]{qureshi2021detecting}
Qureshi, A., Meg{\'\i}as, D., and Kuribayashi, M.
\newblock Detecting deepfake videos using digital watermarking.
\newblock In \emph{2021 Asia-Pacific Signal and Information Processing Association Annual Summit and Conference}, pp.\  1786--1793. IEEE, 2021.

\bibitem[{Qwen Team}(2024)]{qvq}
{Qwen Team}.
\newblock Qvq: To see the world with wisdom, December 2024.
\newblock URL \url{https://qwenlm.github.io/blog/qvq-72b-preview/}.

\bibitem[Shang et~al.(2021)Shang, Kou, Zhang, and Wang]{tiktec}
Shang, L., Kou, Z., Zhang, Y., and Wang, D.
\newblock A multimodal misinformation detector for covid-19 short videos on tiktok.
\newblock In \emph{2021 IEEE international conference on big data (big data)}, pp.\  899--908. IEEE, 2021.

\bibitem[Sheng et~al.(2025)Sheng, Qi, Yang, Bu, Hsu, Lee, and Cao]{sheng2025combating}
Sheng, Q., Qi, P., Yang, T., Bu, Y., Hsu, W., Lee, M.~L., and Cao, J.
\newblock Combating online misinformation videos: Characterization, detection, and prevention.
\newblock In \emph{Proceedings of the 33rd ACM International Conference on Multimedia}, pp.\  14344--14345, 2025.

\bibitem[Sun et~al.(2025)Sun, Jin, Wang, Wang, Ma, Wang, Geng, Wu, Zhang, and Liu]{fast}
Sun, G., Jin, M., Wang, Z., Wang, C.-L., Ma, S., Wang, Q., Geng, T., Wu, Y.~N., Zhang, Y., and Liu, D.
\newblock Visual agents as fast and slow thinkers, 2025.
\newblock URL \url{https://arxiv.org/abs/2408.08862}.

\bibitem[Wang et~al.(2025{\natexlab{a}})Wang, Liu, Zhang, and Wang]{wang2025consistency}
Wang, J., Liu, J., Zhang, N., and Wang, Y.
\newblock Consistency-aware fake videos detection on short video platforms.
\newblock In \emph{International Conference on Intelligent Computing}, pp.\  200--210. Springer, 2025{\natexlab{a}}.

\bibitem[Wang et~al.(2023)Wang, Zhang, Xu, Xu, Xu, and Wang]{wang2023cross}
Wang, L., Zhang, C., Xu, H., Xu, Y., Xu, X., and Wang, S.
\newblock Cross-modal contrastive learning for multimodal fake news detection.
\newblock In \emph{Proceedings of the 31st ACM International Conference on Multimedia}, pp.\  5696--5704, 2023.

\bibitem[Wang et~al.(2024{\natexlab{a}})Wang, Bai, Tan, Wang, Fan, Bai, Chen, Liu, Wang, Ge, Fan, Dang, Du, Ren, Men, Liu, Zhou, Zhou, and Lin]{qwen2}
Wang, P., Bai, S., Tan, S., Wang, S., Fan, Z., Bai, J., Chen, K., Liu, X., Wang, J., Ge, W., Fan, Y., Dang, K., Du, M., Ren, X., Men, R., Liu, D., Zhou, C., Zhou, J., and Lin, J.
\newblock Qwen2-vl: Enhancing vision-language model's perception of the world at any resolution, 2024{\natexlab{a}}.
\newblock URL \url{https://arxiv.org/abs/2409.12191}.

\bibitem[Wang et~al.(2024{\natexlab{b}})Wang, Chen, Wang, Cao, Liu, Gao, Zhu, Zhu, Lu, Qiao, et~al.]{intervl-mpo}
Wang, W., Chen, Z., Wang, W., Cao, Y., Liu, Y., Gao, Z., Zhu, J., Zhu, X., Lu, L., Qiao, Y., et~al.
\newblock Enhancing the reasoning ability of multimodal large language models via mixed preference optimization.
\newblock \emph{arXiv preprint arXiv:2411.10442}, 2024{\natexlab{b}}.

\bibitem[Wang et~al.(2018)Wang, Ma, Jin, Yuan, Xun, Jha, Su, and Gao]{eann}
Wang, Y., Ma, F., Jin, Z., Yuan, Y., Xun, G., Jha, K., Su, L., and Gao, J.
\newblock Eann: Event adversarial neural networks for multi-modal fake news detection.
\newblock In \emph{Proceedings of the 24th acm sigkdd international conference on knowledge discovery \& data mining}, pp.\  849--857, 2018.

\bibitem[Wang et~al.(2025{\natexlab{b}})Wang, Qian, and Li]{fmnv}
Wang, Y., Qian, Z., and Li, P.
\newblock Fmnv: A dataset of media-published news videos for fake news detection.
\newblock In \emph{International Conference on Intelligent Computing}, pp.\  321--332. Springer, 2025{\natexlab{b}}.

\bibitem[Wei et~al.(2022)Wei, Wang, Schuurmans, Bosma, Xia, Chi, Le, Zhou, et~al.]{chain}
Wei, J., Wang, X., Schuurmans, D., Bosma, M., Xia, F., Chi, E., Le, Q.~V., Zhou, D., et~al.
\newblock Chain-of-thought prompting elicits reasoning in large language models.
\newblock \emph{Advances in neural information processing systems}, 35:\penalty0 24824--24837, 2022.

\bibitem[Xu et~al.(2025)Xu, Li, Wang, and Jiang]{xu2025ample}
Xu, X., Li, X., Wang, T., and Jiang, Y.
\newblock Ample: Emotion-aware multimodal fusion prompt learning for fake news detection.
\newblock In \emph{International Conference on Multimedia Modeling}, pp.\  86--100. Springer, 2025.

\bibitem[Yang et~al.(2025)Yang, Li, Yang, Zhang, Hui, Zheng, Yu, Gao, Huang, Lv, Zheng, Liu, Zhou, Huang, Hu, Ge, Wei, Lin, Tang, Yang, Tu, Zhang, Yang, Yang, Zhou, Zhou, Lin, Dang, Bao, Yang, Yu, Deng, Li, Xue, Li, Zhang, Wang, Zhu, Men, Gao, Liu, Luo, Li, Tang, Yin, Ren, Wang, Zhang, Ren, Fan, Su, Zhang, Zhang, Wan, Liu, Wang, Cui, Zhang, Zhou, and Qiu]{qwen3}
Yang, A., Li, A., Yang, B., Zhang, B., Hui, B., Zheng, B., Yu, B., Gao, C., Huang, C., Lv, C., Zheng, C., Liu, D., Zhou, F., Huang, F., Hu, F., Ge, H., Wei, H., Lin, H., Tang, J., Yang, J., Tu, J., Zhang, J., Yang, J., Yang, J., Zhou, J., Zhou, J., Lin, J., Dang, K., Bao, K., Yang, K., Yu, L., Deng, L., Li, M., Xue, M., Li, M., Zhang, P., Wang, P., Zhu, Q., Men, R., Gao, R., Liu, S., Luo, S., Li, T., Tang, T., Yin, W., Ren, X., Wang, X., Zhang, X., Ren, X., Fan, Y., Su, Y., Zhang, Y., Zhang, Y., Wan, Y., Liu, Y., Wang, Z., Cui, Z., Zhang, Z., Zhou, Z., and Qiu, Z.
\newblock Qwen3 technical report, 2025.
\newblock URL \url{https://arxiv.org/abs/2505.09388}.

\bibitem[Yang et~al.(2023)Yang, Li, Lin, Wang, Lin, Liu, and Wang]{gpt4v}
Yang, Z., Li, L., Lin, K., Wang, J., Lin, C.-C., Liu, Z., and Wang, L.
\newblock The dawn of lmms: Preliminary explorations with gpt-4v(ision), 2023.
\newblock URL \url{https://arxiv.org/abs/2309.17421}.

\bibitem[Yuan et~al.(2025)Yuan, Qu, Qian, Chen, Tang, Sun, Chu, Zhang, Wang, Cai, and Li]{video-star}
Yuan, Z., Qu, X., Qian, C., Chen, R., Tang, J., Sun, L., Chu, X., Zhang, D., Wang, Y., Cai, Y., and Li, S.
\newblock Video-star: Reinforcing open-vocabulary action recognition with tools, 2025.
\newblock URL \url{https://arxiv.org/abs/2510.08480}.

\bibitem[Zhang et~al.(2025{\natexlab{a}})Zhang, Li, Zhang, Chen, Liu, Lin, Yan, Liu, and Zha]{fact-r1}
Zhang, F., Li, D., Zhang, Q., Chen, J., Liu, G., Lin, J., Yan, J., Liu, J., and Zha, Z.-J.
\newblock Fact-r1: Towards explainable video misinformation detection with deep reasoning, 2025{\natexlab{a}}.
\newblock URL \url{https://arxiv.org/abs/2505.16836}.

\bibitem[Zhang et~al.(2025{\natexlab{b}})Zhang, Fang, Yang, and Feng]{Llava-mini}
Zhang, S., Fang, Q., Yang, Z., and Feng, Y.
\newblock Llava-mini: Efficient image and video large multimodal models with one vision token, 2025{\natexlab{b}}.
\newblock URL \url{https://arxiv.org/abs/2501.03895}.

\bibitem[Zhao et~al.(2025)Zhao, Zhang, Lin, Li, Wu, Zhang, and Wei]{pyvision}
Zhao, S., Zhang, H., Lin, S., Li, M., Wu, Q., Zhang, K., and Wei, C.
\newblock Pyvision: Agentic vision with dynamic tooling, 2025.
\newblock URL \url{https://arxiv.org/abs/2507.07998}.

\bibitem[Zheng et~al.(2025)Zheng, Yang, Hong, Zhao, Xu, Yang, Shen, and Yu]{zheng2025deepeyes}
Zheng, Z., Yang, M., Hong, J., Zhao, C., Xu, G., Yang, L., Shen, C., and Yu, X.
\newblock Deepeyes: Incentivizing" thinking with images" via reinforcement learning.
\newblock \emph{arXiv preprint arXiv:2505.14362}, 2025.

\end{thebibliography}
\bibliographystyle{icml2026}

\newpage
\appendix
\onecolumn
\section*{Appendix}
\label{sec:app}
In this appendix, we provide more experimental details, related work, tool library, and discussions
for a comprehensive evaluation and understanding of our method. Detailed contents are as follows:
\section{Experiment}
\subsection{More Experimental Details}
\label{sub:exdetail}
\paragraph{Datasets.} A variety of datasets have been developed to facilitate research on video-based misinformation detection. Early datasets typically focused on narrow domains, such as medical misinformation~\cite{hou2019towards} or COVID-19-related content~\cite{knuutila2021dataset}, and often covered multiple languages. While valuable for domain-specific analysis, these datasets are generally limited in scale, topical diversity, and long-term public availability.

More recent benchmarks have shifted toward large-scale, short-video misinformation detection. In particular, FakeSV~\cite{fakesv} consists of short news-style videos collected from social media platforms, paired with textual metadata and crowd-sourced annotations. FakeTT~\cite{fakett} focuses on TikTok videos and incorporates multimodal signals including visual content, audio transcripts, and user interaction features. FakeVV~\cite{fact-r1} focuses on more complex and ambiguous misinformation scenarios and offers richer multimodal signals and annotations, enabling more comprehensive evaluation of reasoning-oriented verification models.

Together, these datasets provide complementary coverage in terms of content sources, multimodal structure, and verification difficulty, enabling comprehensive evaluation of video misinformation detection systems under diverse real-world conditions. Accordingly, all experimental evaluations in this work are conducted on FakeSV, FakeTT, and FakeVV.
\paragraph{Training Details.} To support agentic reasoning and tool-aware decision making, we first construct a multimodal agentic Chain-of-Thought (CoT) dataset tailored for video misinformation detection. Starting from the training sets of FakeSV, FakeTT, and FakeVV, we employ a stronger teacher model, Qwen2.5-VL-72B, to generate agentic reasoning trajectories under carefully designed prompts. These prompts explicitly encourage uncertainty assessment, tool-selection intent,
evidence grounding, and final verification decisions conditioned on multimodal inputs.

Given the multimodal input and the ground-truth label, the teacher model produces full agent
trajectories, including intermediate reasoning steps, tool invocation decisions, and
evidence-aware conclusions. To ensure data quality, we apply a two-stage filtering process.
First, rule-based validation removes malformed trajectories, such as missing reasoning structure,
invalid tool actions, or incorrect final decisions. Second, a subset of samples is manually
inspected to further eliminate low-quality or hallucinated reasoning traces. Only high-quality,
coherent, and evidence-grounded trajectories are retained for supervised fine-tuning.

We perform agentic supervised fine-tuning (SFT) on Qwen2.5-VL-7B-Instruct using the curated
agentic CoT dataset. The objective of this stage is to inject structured reasoning patterns,
explicit uncertainty awareness, and tool-invocation behaviors into the base model prior to
reinforcement learning.

Training is conducted with mixed-precision (BF16) and DeepSpeed optimization. We use a per-device batch size of 1 with gradient accumulation and train the model for one epoch. Gradient checkpointing and FlashAttention are enabled to reduce memory consumption. This SFT stage provides essential inductive bias for evidence-seeking and multi-step reasoning, which we find to be critical for stabilizing and guiding subsequent reinforcement learning.

Following supervised fine-tuning, we further optimize FactGuard using decision-aware reinforcement learning with Group Relative Policy Optimization (GRPO). All reinforcement learning experiments are conducted on 8 NVIDIA H100 GPUs.

During GRPO training, each input prompt is unfolded into 8 candidate trajectories, capturing diverse reasoning paths, tool-invocation behaviors, and final verification outcomes. The video-clip inspection tool is restricted to at most one invocation per prompt, encouraging selective and deliberate visual evidence acquisition, while external knowledge retrieval remains unconstrained. Unless otherwise specified, the model is configured with a balanced cost setting ($\alpha:\gamma = 1:1$), encouraging neutral treatment of false positives and false negatives during trajectory generation.

We set the learning rate to $1\times10^{-6}$ and train the policy for one epoch. The maximum prompt length is set to 16{,}384 tokens and the maximum response length is 768 tokens. The KL regularization coefficient is fixed to 0.04 to stabilize policy updates against a frozen reference model. We use a per-device batch size of 1 without gradient accumulation.

This reinforcement learning stage refines verification decisions under task-aligned rewards, explicitly calibrating tool usage and asymmetric error costs, and further strengthens evidence-grounded reasoning beyond supervised imitation.
\subsection{More Results}
\begin{figure*}[t]
  \centering
  \includegraphics[width=1\linewidth]{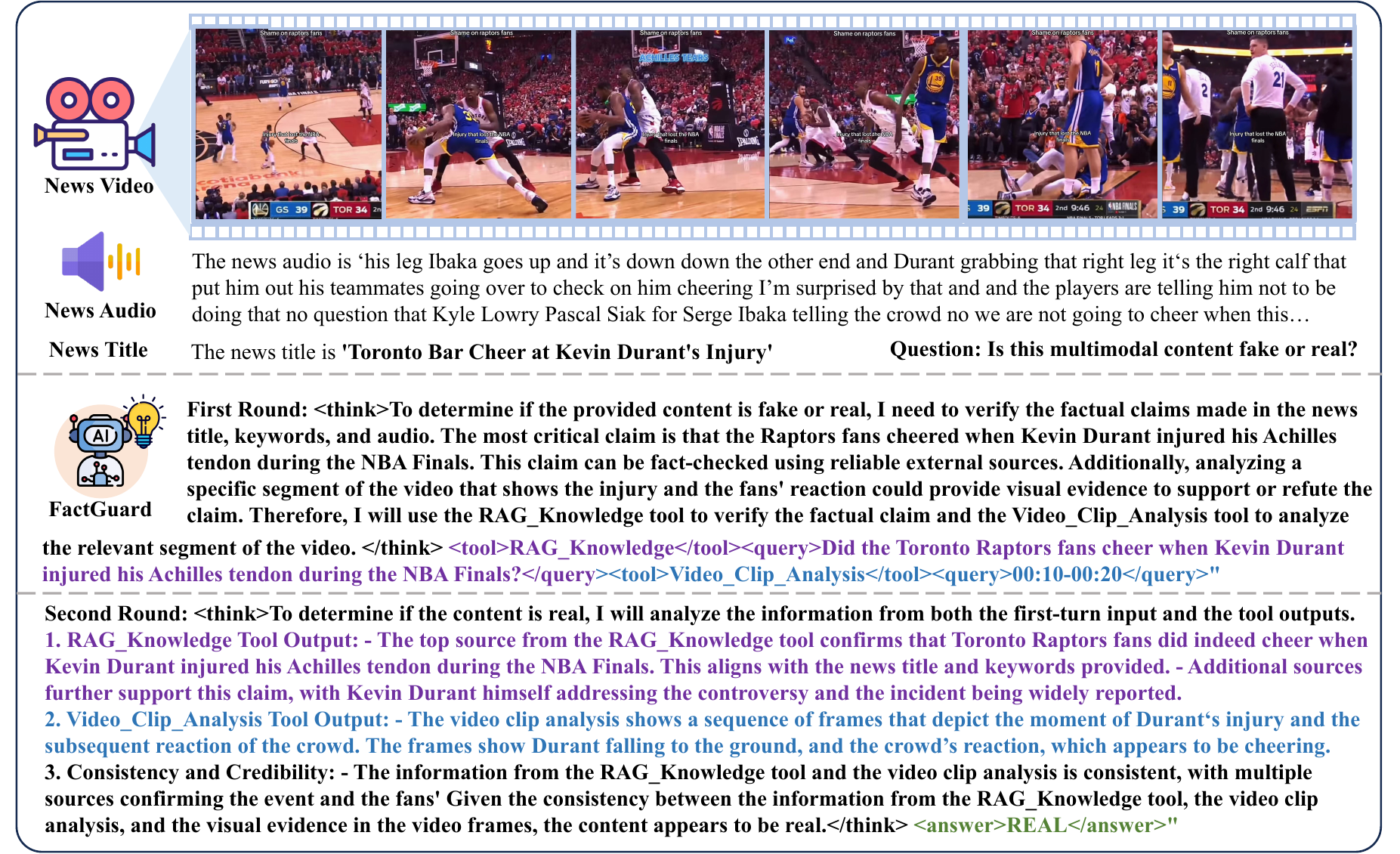}
  \caption{
Additional Case Study of FactGuard.
}
  \label{fig:durant}
\end{figure*}

As illustrated in Figure~\ref{fig:durant}, when confronted with uncertain or insufficient evidence, FactGuard adaptively invokes multiple tools to refine its verification process.
Specifically, FactGuard first calls \textit{FactProbe} to retrieve external knowledge and verify the factual validity of the claimed event.
It then employs \textit{ClipScout} to perform focused visual analysis, attending to the reactions of surrounding spectators in the video.
By integrating the retrieved factual evidence with the localized visual cues, FactGuard is able to conduct a more informed and reliable reasoning process, ultimately leading to a more accurate and well-supported conclusion.
\section{Tool Library Details}
FactGuard is equipped with a lightweight yet effective tool library to support evidence acquisition during agentic verification. The tool library is designed to complement the model’s internal multimodal reasoning by providing access to external factual knowledge and localized visual evidence. In this work, we implement two core tools: an external knowledge retrieval tool (\textbf{FactProbe}) and a video clip inspection tool (\textbf{ClipScout}). These tools are selectively invoked by the agent when internal reasoning alone is insufficient for confident verification.

\subsection{External Knowledge Retrieval (FactProbe)}
FactProbe implements a retrieval-augmented generation (RAG) pipeline for external factual verification. Given a structured factual query produced by the agent during the first-stage reasoning, the tool issues a web search request via a commercial search API (Serper) to retrieve a small set of relevant results. To reduce noise, only organic search results are retained, and user-generated or social media sources are excluded.

For each query, the tool collects the top-ranked results and extracts their titles, snippets, and source links. The retrieved content is then aggregated into a compact textual report that summarizes the external evidence relevant to the queried claim. Rather than performing full document reading, FactProbe adopts a lightweight retrieval-and-snippet strategy to balance evidence coverage and efficiency. The resulting report is passed to the second-stage reasoning process as auxiliary textual input, enabling the model to incorporate externally grounded information during verification.

\subsection{Video Clip Temporal Inspection (ClipScout)}
ClipScout provides targeted temporal inspection of video content by extracting representative frames from a specified time interval. When the agent invokes ClipScout, it outputs a precise temporal query (e.g., a start–end timestamp) corresponding to the most evidence-bearing segment of the video.

The tool decodes the video stream and uniformly samples a small number of frames (four by default) within the queried interval. These frames are then arranged into a $2\times2$ visual grid to form a compact visual summary. To control computational and token overhead, the grid is resized such that its maximum spatial dimension does not exceed a fixed resolution threshold.

The resulting composite image is appended to the multimodal input during the refinement stage, allowing the model to reason over localized visual evidence without processing the entire video. ClipScout is constrained to at most one invocation per instance, ensuring that temporal inspection remains selective and aligned with deliberate evidence acquisition rather than exhaustive visual scanning.

\section{Prompt Templates}

We present the prompt templates used in our framework, including the two-stage prompting strategy for FactGuard and the auditing prompt employed by GPT-4o. FactGuard operates in a structured two-turn manner: the first turn determines whether external evidence is required and selects appropriate tools, while the second turn produces the final verification result conditioned on the gathered evidence. In addition, GPT-4o is used as an external auditor to assess the evidence grounding of the generated reasoning.

\begin{figure*}[t]
  \centering
  \includegraphics[width=1\linewidth]{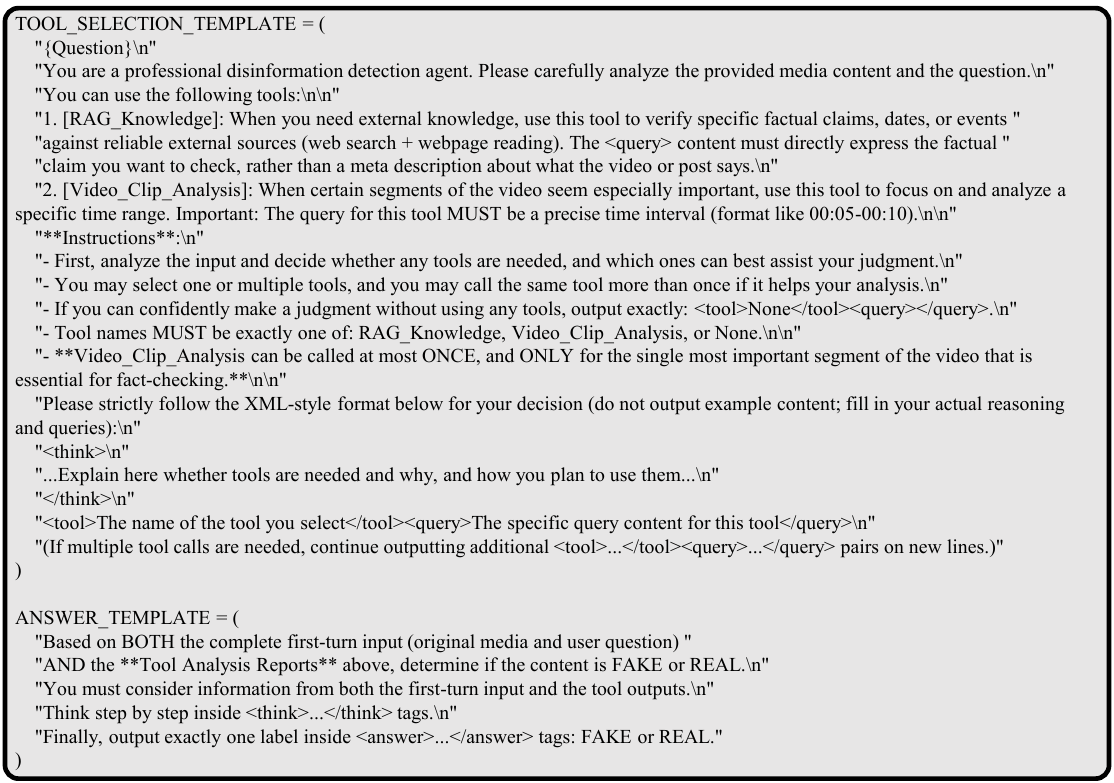}
  \caption{
  \textbf{FactGuard Two-Turn Prompting.}
  }
  \label{fig:factguard_turn}
\end{figure*}

\paragraph{FactGuard's Two-Turn Prompting.}
As shown in Figure~\ref{fig:factguard_turn}, FactGuard adopts a two-turn prompting strategy to enable explicit evidence-seeking and grounded verification. In the first turn, the model analyzes the input claim and decides whether external tools should be invoked to acquire additional evidence. In the second turn, FactGuard integrates the retrieved evidence with its internal reasoning to produce a final prediction and explanation. This design explicitly separates evidence acquisition from answer generation, reducing reliance on unsupported internal assumptions.

\begin{figure*}[t]
  \centering
  \includegraphics[width=1\linewidth]{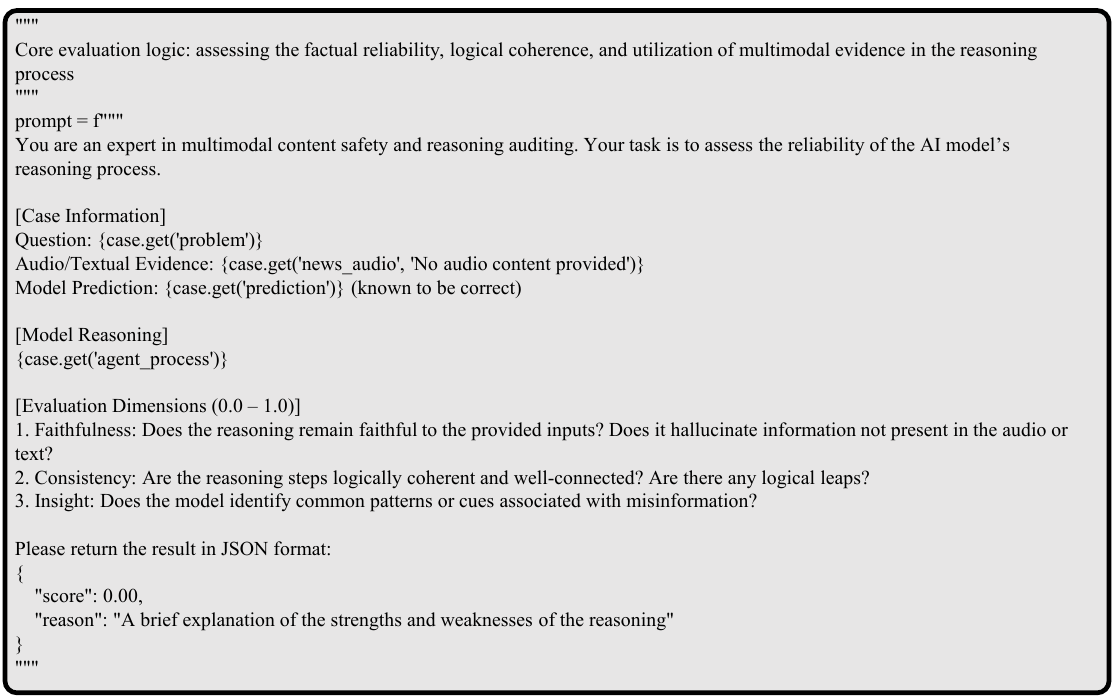}
  \caption{
  \textbf{GPT-4o reasoning audit prompt.}
  The prompt evaluates whether a model’s reasoning and prediction are supported by explicit and relevant evidence, serving as an external assessment of reasoning faithfulness.
  }
  \label{fig:gpt4o_audit}
\end{figure*}

\paragraph{GPT-4o's Reasoning Audit Prompt.}
To assess the faithfulness, logical consistency, and evidence grounding of model-generated reasoning, we adopt GPT-4o as an external auditor with a dedicated evaluation prompt. Given the model’s prediction, reasoning trace, and associated evidence, GPT-4o evaluates whether the conclusion is logically coherent, internally consistent, and supported by explicit evidence rather than speculative or internally generated assumptions, as illustrated in Figure~\ref{fig:gpt4o_audit}. This auditing procedure provides an additional layer of quality control to evaluate theity and interpretinterpretability of reasoningning.


\end{document}